%% file: main.tex
\documentclass[conference]{IEEEtran}
\usepackage{times}

\usepackage{graphicx}	
\usepackage[numbers]{natbib}
\usepackage{multicol}
\usepackage{amsfonts}
\usepackage{amsmath}
\usepackage{booktabs}
\usepackage[bookmarks=true]{hyperref}
\usepackage{multirow}
\usepackage{subcaption}
\usepackage[table]{xcolor}
\definecolor{softviolet}{RGB}{230,220,240}

\begin{document}

\title{
PointACT: Vision-Language-Action Models with Multi-Scale Point-Action Interaction
}

\author{Shizhe Chen, Paul Pacaud, Cordelia Schmid\\
Inria, \'Ecole normale sup\'erieure, CNRS, PSL Research University\\
\url{https://cshizhe.github.io/projects/pointact.html}
}

\maketitle

\input{sec/00_abstract}

\IEEEpeerreviewmaketitle

\input{sec/01_intro}

\input{sec/02_related}

\input{sec/03_method}

\input{sec/04_expr}

\input{sec/10_conclusion}

\section*{Acknowledgments}

{
\footnotesize
This work was granted access to HPC resources of IDRIS under the allocation AD011014846 made by GENCI. 
It was funded in part by the French government under management of Agence Nationale de la Recherche as part of the “France 2030" program, reference ANR-23-IACL-0008 (PR[AI]RIE-PSAI projet), the ANR project 3D-GEM (ANR-25-CE23-7777-01), the ANR project VideoPredict (ANR-21-FAI1-0002-01) and the Paris Île-de-France Région in the frame of the DIM AI4IDF. 
Cordelia Schmid would like to acknowledge the support by the Körber European Science Prize.
}

\bibliographystyle{plainnat}
\bibliography{sec/11_references}

\include{sec/12_appendix}

\end{document}

%% file: sec/00_abstract.tex
\begin{abstract}

Vision–Language–Action (VLA) models have shown strong potential for general-purpose robotic manipulation by leveraging large pretrained vision-language backbones. 
However, most existing VLAs rely primarily on 2D visual representations, which limits their ability to reason about fine-grained geometry and spatial grounding - capabilities that are essential for precise and robust manipulation in 3D environments.
In this paper, we propose PointACT, a dual-system 3D-aware VLA policy that integrates hierarchical 3D point cloud representations directly into the action decoding process. 
PointACT employs a multi-scale point-action interaction mechanism with efficient bottleneck window self-attention, enabling evolving action tokens to densely attend to both local geometric detail and global scene structure.  
We evaluate PointACT on the LIBERO and RLBench benchmarks and systematically compare it against monolithic and dual-system VLA baselines, including variants augmented with point cloud inputs. 
PointACT achieves consistent improvements across both benchmarks, increasing success rates by 10\% on the challenging RLBench-10Tasks suite over state-of-the-art pretrained VLAs, with even larger gains when the vision–language backbone is frozen and the action expert is trained from scratch.
Extensive ablation studies demonstrate that tightly coupling hierarchical 3D geometry with pretrained 2D semantic representations is critical for robust and spatially grounded robot control.
Our results also highlight the promise of pretrained 3D representations for 3D-aware VLA policies.

\end{abstract}

%% file: sec/01_intro.tex
\section{Introduction}
\label{sec:intro}

Developing general-purpose robots that can map natural language instructions to diverse physical actions remains a central challenge in robotics. 
Recently, Vision–Language–Action models (VLAs)~\cite{kim2025openvla,black2024pi_0,bjorck2025gr00t,chen2025internvlam1,qu2025eo1} have emerged as a promising paradigm by building robotic control policies on top of large pretrained vision–language models (VLMs)~\cite{achiam2023gpt4,bai2025qwen2_5vl}. 
Hence, VLAs inherit a strong semantic understanding of the world from VLMs, enabling broader generalization across objects, scenes, and instructions than traditional task-specific policies.

However, a fundamental mismatch remains: the physical world is inherently three-dimensional, while most state-of-the-art VLAs rely on 2D image representations. 
The absence of explicit geometric input weakens spatial understanding and limits performance in precise and robust manipulation. 
As a result, current VLAs often depend on large-scale data to achieve viewpoint invariance and still struggle with tasks that require accurate spatial reasoning~\cite{wu2024robomind}. 
Recent VLA approaches~\cite{qu2025spatialvla, nikolov2025spear1,lee2025molmoact,yuan2025depthvla} attempt to recover 3D structure from single-view or sparse-view RGB images, either using external depth models~\cite{bhat2023zoedepth} or jointly trained predictors, but such reconstruction remains inherently ambiguous and unreliable for high-accuracy control.

With the growing availability of 3D sensors such as RGB-D cameras, explicit geometric observations are increasingly accessible, yet how to effectively integrate them into VLAs remains an open challenge. Some 3D-aware VLAs incorporate depth or 3D positional cues as auxiliary signals layered onto 2D feature representations~\cite{li20253dsvla,zhen20243dvla}, where action generation is still primarily driven by image tokens and geometric structure is weakly coupled to the final control output. 
Other approaches introduce 3D inputs into action experts~\cite{li2026pointvla,sun2025geovla, yang2025fp3}, but typically rely on coarse-grained point cloud features, limiting fine-grained geometric interaction. 
Moreover, prior studies report limited gains from transferring pretrained 3D representations to policy learning~\cite{li2026pointvla,ze20243ddiffusionpolicy}, partly due to the limited availability of in-domain pretrained point cloud encoders.

In this work, we propose \textbf{PointACT}, a 3D-enhanced VLA framework that combines a pretrained VLM backbone and a 3D action expert to integrate point clouds into action generation. 
Geometric observations are encoded using a strong point-cloud encoder based on pretrained Point Transformer v3~\cite{wu2024pctv3}, which provides useful geometric features despite being trained on out-of-domain data.
We systematically investigate different strategies for incorporating 3D information into VLAs, including injecting point tokens into the VLM backbone in monolithic VLAs and integrating geometry with action experts in dual-system VLA architectures, as illustrated in Figure~\ref{fig:teaser}(c).
To foster action prediction grounded in 3D, we further propose a point-action expert that enables multi-scale interaction between point tokens and action tokens, allowing geometric cues to continuously condition the evolving action tokens while capturing both global structure and local detail. 
To ensure efficiency, action tokens serve as bottleneck queries that attend to point tokens through windowed attention, avoiding the computational cost of dense point token interactions.

We evaluate PointACT on the LIBERO and RLBench simulators across diverse tasks and action types, including keypoint pose prediction and delta action movements. 
Ablation studies show that injecting point tokens into the VLM backbone is less effective, as it can interfere with pretrained representations. 
In contrast, integrating point clouds within the action expert, particularly through fine-grained point–action interaction, produces consistent and substantial gains. 
We further find that initializing the point-cloud expert with a pretrained point encoder provides useful geometric priors and improves performance, despite domain gaps between large-scale 3D pretraining data and tabletop manipulation scenes. 

\input{figs_tex/teaser}

Our contributions in this work are threefold:
\begin{itemize}
    \item We systematically study strategies to improve 3D spatial understanding of VLAs with point clouds, comparing backbone injection vs. action-expert integration.
    
    \item We propose PointACT, a dual-system VLA with a pretrained VLM backbone and a dedicated point-action expert that enables fine-grained point–action interaction through an efficient bottleneck windowed self-attention over multi-scale point cloud features.
    
    \item PointACT achieves consistent gains on LIBERO and RLBench benchmarks across action types, demonstrating the benefit of fine-grained geometry-driven action prediction.
    
\end{itemize}

%% file: figs_tex/teaser.tex
\begin{figure*}[thbp]
    \centering
    \begin{subfigure}{0.29\textwidth}
        \centering
        \includegraphics[width=\linewidth]{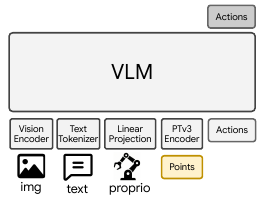}
        \caption{Monolithic 3D-aware VLA}
        \label{fig:teaser_monolithic_3dvla}
    \end{subfigure}
    \hfill
    \begin{subfigure}{0.31\textwidth}
        \centering
        \includegraphics[width=\linewidth]{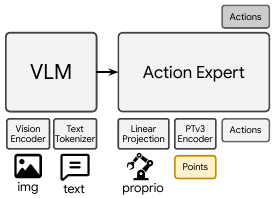}
        \caption{Dual-system 3D-aware VLA}
        \label{fig:teaser_2system_3dvla}
    \end{subfigure}
    \hfill
    \begin{subfigure}{0.33\textwidth}
        \centering
        \includegraphics[width=\linewidth]{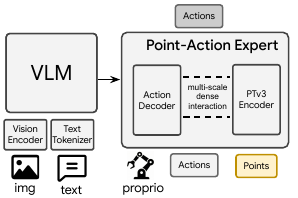}
        \caption{The proposed PointACT VLA}
        \label{fig:teaser_pointact}
    \end{subfigure}
    \caption{
    \textbf{Comparison of 3D integration strategies in VLAs.}
(a) Monolithic 3D-aware VLA: 3D point features are fed directly into the pretrained VLM backbone, which largely increases the computation burden and may disrupt pretrained representations.
(b) Dual-system 3D-aware VLA: 3D information is introduced into a separate action expert, but typically through coarse-grained global features with limited interaction between geometry and actions.
(c) PointACT (ours): a dedicated point–action expert enables fine-grained, multi-scale interaction between point cloud features and action tokens during decoding, preserving the pretrained VLM while allowing geometry to directly shape action generation.
}
\label{fig:teaser}
\end{figure*}

%% file: sec/02_related.tex
\section{Related Work}
\label{sec:related}

\subsection{Learning-based Visuomotor Policies}

Learning-based visuomotor policies have advanced rapidly with strong neural network architectures~\cite{zhao2023act,jang2022bcz,brohan2022rt1} and training objectives~\cite{chi2025diffusionpolicy,pan2025muchado}.
While 2D visual policies benefit from powerful pretrained representations~\cite{radford2021clip,simeoni2025dinov3}, they often struggle with precise spatial reasoning due to the lack of explicit geometry grounding~\cite{chen2024spatialvlm}. 
Depth-based methods partially address this limitation for grasping and planning~\cite{mahler2017dexnet2,ait2023simultaneous}, but depth remains a 2.5D representation.
To enable richer spatial reasoning, recent works explore explicit 3D representations such as voxels~\cite{james2022c2farm,shridhar2023peract} and point clouds~\cite{liu2023framemining,chen2023polarnet,garcia2025gembench,ze20243ddiffusionpolicy,chen2025vividex} for robotic manipulation.
SUGAR~\cite{chen2024sugar} further strengthens point-based policies through 3D representation pretraining with simulated data.
Hybrid approaches fuse the semantic strength of 2D features with the geometric accuracy of 3D reasoning. HiveFormer~\cite{guhur2023hiveformer} and RVT~\cite{goyal2023rvt,goyal2024rvt2} predict multiview 2D action heatmaps and lift them to 3D coordinates using depth, while Act3D~\cite{gervet2023act3d} and 3D Diffuser Actor~\cite{ke20243ddifusseractor} lift pretrained 2D features into 3D volumes and learn fully 3D action policies. 
However, these visuomotor policies are typically trained and evaluated on narrow task suites, limiting  generalization.

\subsection{Vision-Language-Action Models (VLAs)}
The success of LLMs~\cite{achiam2023gpt4} and VLMs~\cite{bai2025qwen2_5vl} has inspired their use in robotic manipulation.
Early works have explored zero-shot application of these foundation models for high-level planning~\cite{huang2023voxposer}, providing motion constraints~\cite{huang2024copa}, or code generation~\cite{liang2023code}, but the performance remained limited without training on robot-specific datasets.

This motivates end-to-end VLAs which extend VLMs by adding an action modality and fine-tuning on robot datasets~\cite{o2024openx,bu2025agibot}.
RT-2~\cite{zitkovich2023rt2} and OpenVLA~\cite{kim2025openvla} use autoregressive next token prediction with discretized action tokens;
Octo~\cite{ghosh2024octo} and EO1~\cite{qu2025eo1} introduce continuous action heads with diffusion training objectives~\cite{chi2025diffusionpolicy};
OpenVLA-OFT~\cite{kim2025openvlaoft} further studies VLA design and achieves strong performance and efficiency via simple regression.
To enhance reasoning, several works incorporate Chain-of-Thought (CoT)~\cite{wei2022cot}. 
ECoT~\cite{zawalski2025ecot} performs multi-step reasoning about plans, sub-tasks, motions, and visual grounding before predicting actions, while CoT-VLA~\cite{zhao2025cotvla} generates future image frames as visual goals.
ThinkAct~\cite{huang2025thinkact} further optimizes the reasoning process via reinforcement learning.
However, relying on a single monolithic VLM for reasoning and control has raised concerns about inefficiency and catastrophic forgetting. 
Therefore, recent methods~\cite{black2024pi_0,intelligence2504pi05,bjorck2025gr00t,shukor2025smolvla,zheng2025xvla,chen2025internvlam1,luo2026beingh05} such as $\pi_0$~\cite{black2024pi_0} and GR00T~\cite{bjorck2025gr00t} adopt a dual-system framework that decouples high-level perception and planning from low-level control. 
In these systems, the VLM is typically kept frozen, and its output representations are fed into a lightweight action expert for continuous action generation.
Despite strong semantic capabilities, existing VLAs remain largely 2D-centric and are limited in 3D understanding and spatial generalization~\cite{zhou2025liberopro,fei2025liberoplus,pumacay2024colosseum}. In this work, we address this by equipping VLAs with a 3D action expert for more accurate 3D action prediction.

\subsection{3D Spatial Reasoning in VLAs}

Recent VLAs increasingly incorporate 3D cues to improve spatial reasoning and generalization. 
One line of work augments RGB observations with predicted depth, such as fusing image and depth features within a monolithic VLA~\cite{qu2025spatialvla, nikolov2025spear1}.
To alleviate the reliance on external depth models, MolmoAct~\cite{lee2025molmoact} generates depth tokens inside the VLM at the cost of token-generation overhead. 
DepthVLA~\cite{yuan2025depthvla} more tightly couples depth with control by integrating a depth transformer into the action expert and finetunes the depth model.

Another line of work directly incorporates sensor-derived geometry such as RGB-D or point clouds. 
Some works~\cite{zhen20243dvla,li20253dsvla} fuse point clouds and image representations as inputs to the VLM, which often requires training on large-scale 3D datasets.
To better leverage pretrained 2D VLMs, BridgeVLA~\cite{li2025bridgevla} relies on multi-view images to predict 2D action heatmaps and lifts them to 3D using camera poses and depth. While effective for long-horizon keypoint pose prediction, this design is specialized for positional control and is difficult to extend to other action formats like delta movements or velocity control.
Other point cloud augmented VLAs~\cite{li2026pointvla, sun2025geovla} train point cloud encoders to inject the learned 3D embeddings into action experts. However, these methods primarily rely on coarse-grained global geometry features, and observe no benefit from pretraining the point cloud encoder~\cite{li2026pointvla}.
Our method instead integrates point clouds into the action expert at multiple scales, enabling dense interactions between 3D point features and action tokens, and demonstrates strong performance across multiple benchmarks and action types.

%% file: sec/03_method.tex
\section{Method}
\label{sec:method}

\subsection{Problem Formulation}

We study vision-and-language guided policy learning for robotic manipulation. 
At each timestep $t$, the agent receives a multi-modal observation $O_t=\{I_t, P_t, s_t, L\}$.
Here, $I_t \in \mathbb{R}^{N_I \times H_I \times W_I \times 3}$ represents multi-view RGB images, $L$ is a natural language instruction, and $s_t$ is the robot's proprioceptive state.
To provide explicit geometric grounding, we include a colored 3D point cloud $P_t \in \mathbb{R}^{N_P \times 6}$.
We do not assume that $P_t$ is pre-aligned with $I_t$, although such alignment can be recovered given calibrated camera poses or multi-view geometric reconstruction.

The objective is to learn a policy $\pi_{\theta}$ that predicts a sequence of $H$ future actions $A_t = (a_t, \ldots, a_{t+H-1})$ to accomplish the task specified by $L$:
\begin{equation}
    A_t = \pi_{\theta}(I_t, P_t, s_t, L),
\end{equation}
where each $a_t$ may represent end-effector pose, or joint-space commands depending on the control interface. 

\subsection{Preliminary: 3D-aware VLAs}
\label{sec:method_baselines}

Our goal is to enhance standard 2D-centric VLA policies with explicit 3D geometric reasoning. 
To contextualize our proposed model, we present two baseline strategies for injecting 3D information into VLA models: a \emph{monolithic} integration strategy and a \emph{dual-system} modular strategy (Figure~\ref{fig:teaser}~(a),(b)). 
We first describe the point cloud encoder shared by both strategies and then present them in detail. 
We conclude by pointing out the limitations of both strategies and how we improve over them.

\noindent\textbf{Point Cloud Encoder.}
To encode geometric observations $P_t$, we employ Point Transformer v3 (PTv3)~\cite{wu2024pctv3}. PTv3 operates directly on unordered point sets and uses a serialization-based grouping strategy to construct local neighborhoods. Self-attention is performed within each group, enabling efficient computation while preserving local geometric structure.
The network follows a hierarchical design in which point features are progressively downsampled across stages, producing multi-scale geometric representations that capture both local detail and global layout.
We initialize the encoder from a PTv3 model pretrained via large-scale self-supervised learning on building-level 3D scenes~\cite{zhang2025concerto}. Although this pretraining domain differs from manipulation settings, it provides strong low-level geometric priors.

\noindent\textbf{Monolithic 3D-aware VLA.}
This approach treats geometric data as an additional input modality to a unified VLM backbone (Figure~\ref{fig:teaser_monolithic_3dvla}).
We build upon the EO1~\cite{qu2025eo1} VLA architecture, which utilizes Qwen2.5-VL~\cite{bai2025qwen2_5vl} as the foundation.
The multi-view images are processed by a pretrained vision encoder, while text tokens and proprioceptive state $s_t$ are mapped into the VLM's latent space via an embedding layer and a linear projection, respectively.
We further project point embeddings from the final layer of the PTv3 encoder into the same latent space, and concatenate all the modality tokens as inputs to the VLM.
This enables the method to perform joint conditioning on image, language, and geometry within a single transformer.
The model is trained using a flow-matching objective~\cite{lipman2023flowmatching}, where the VLM receives noised action tokens and learns to predict the velocity field for iterative denoising.

\noindent\textbf{Dual-system 3D-aware VLA.}
This approach decouples semantic reasoning from motor control (Figure~\ref{fig:teaser_2system_3dvla}), following the GR00T~\cite{bjorck2025gr00t} VLA architecture.
However, instead of using the GR00T pretrained weights, we use the same Qwen2.5-VL as the VLM backbone for fair comparison with the monolithic VLA variant.
The VLM backbone remains frozen to preserve its general-purpose representations.
A separate \emph{action expert} (e.g., diffusion transformer~\cite{peebles2023dit}) is trained from scratch to predict $A_t$ conditioned on the VLM features, robot's proprioceptive state $s_t$, and point embeddings from the last layer of the PTv3 encoder.

In both paradigms, point features are fused only at a high level, limiting fine-grained geometric influence during action prediction.
We address this limitation with a multi-scale point-action interaction mechanism.

\input{figs_tex/method}

\subsection{PointACT: Multi-scale Point-Action Interaction}

We introduce \textbf{PointACT}, a geometry-aware action expert that injects structured point cloud information into action decoding process through efficient multi-scale point-action interaction. 
Figure~\ref{fig:method} illustrates our model architecture.

Let $Z_{p}^{l} \in \mathbb{R}^{N_p^l \times D^l}$ denote the point embeddings from the $l$-th layer of the PTv3 encoder, where $N_p^l$ is large in early layers and decreases with depth due to hierarchical pooling.
Let $Z_a^l \in \mathbb{R}^{(H+1) \times D^l}$ denote the action token embeddings and the robot proprioceptive state at the corresponding decoding layer, and $Z_{vlm}$ the output embeddings from the VLM backbone.

As $N_p^l$ can be large, we propose a \emph{bottleneck window-based self-attention} mechanism inspired by~\cite{nagrani2021attentionbottlenectk}.
Specifically, we partition the point cloud into $K$ disjoint spatial windows $(W_1, \cdots, W_K)$ using the same serialization strategy as PTv3, with $[N_p^l/K]$ points per window.
For each window $k$, we broadcast the action tokens to the point cloud tokens:
\begin{equation}
    X_k^l = [Z_p^{l, W_k}; Z^l_a].
\end{equation}
We then apply a shared self-attention block across all windows to update the joint point-action representation:
\begin{equation}
    \hat{X}_k^l = \text{Self-Attn}(X_k^l).
\end{equation}
The action tokens serve as a latent bottleneck that aggregates local geometric context, enabling interaction with many local regions while keeping attention complexity linear in the number of windows.
To produce the updated global action representation, we extract the action features from each window and perform an average pool across all $K$ groups:
\begin{equation}
    \hat{Z}_a^l = \frac{1}{K} \sum_{k=1}^K \hat{X}_{k,a}^l
\end{equation}
To distinguish action and point modalities, we employ distinct feed-forward networks (FFN) for action and point tokens after the self-attention.

Following this geometric fusion, the action tokens undergo cross-attention with the VLM embeddings $Z_{vlm}$ to incorporate high-level visual and linguistic instructions:
\begin{equation}
    \bar{Z}_a^l = \text{Cross-Attn}(\hat{Z}_a^l, Z_{vlm}).
\end{equation}
By repeating this process across the hierarchical stages of the PTv3 encoder, the action tokens are conditioned on a spectrum of geometric information, from fine-grained local details to coarse-grained global scene structures.

\subsection{Action Prediction and Training}

PointACT is trained using behavior cloning on demonstration trajectories. 
Depending on the action space, we utilize either a regression-based or a classification-based action head.

\noindent\textbf{Regression Head.}
For continuous control and short-horizon action chunking, we employ a multi-layer perceptron (MLP) to map the output action token embeddings to a sequence of control parameters. 
Following recent findings that regression can be more training-efficient than diffusion-based methods~\cite{kim2025openvlaoft,pan2025muchado}, we optimize the model using an $L_2$ loss over the predicted action chunk per training example:
\begin{equation}
\mathcal{L}_{\text{reg}} 
= \frac{1}{H} 
\sum_{i=t}^{t+H-1} 
\left\| \mathbf{a}_{i} - \mathbf{a}^{*}_{i} \right\|_2^2
\end{equation}
where $a^*_i$ is the ground-truth action.

\noindent\textbf{Classification Head.}
For predicting keypoint actions (e.g., long-term end-effector poses)~\cite{sundaresan2023kite,guhur2023hiveformer}, we adopt a point-anchored classification strategy to better handle the multi-modal distribution of robot trajectories. 
We discretize the target workspace into bins relative to the observed point cloud, as the keypoint often denotes or is close to contact points of the object. Specifically, for a target position $v^* = (x, y, z)$, we compute its coordinates relative to the nearest point $p_j \in P_t$ and assign it to a discrete spatial bin around the point. The rotations are also discretized into $N_{rot}$ bins.

To strengthen geometric grounding, we concatenate the final point embeddings with the action tokens before the prediction layer. The model is trained using a cross-entropy loss:
\begin{equation}
\mathcal{L}_{\text{cls}} = -\sum_{k \in \text\{{pos, rot, open\}}} \sum_{b=1}^{B_k} y_{k,b} \log(\hat{y}_{k,b})
\end{equation}
where $B_k$ denotes the number of bins for each action component. 
This approach allows the model to leverage the point cloud as a spatial anchor for localization.

%% file: figs_tex/method.tex
\begin{figure*}[t]
    \centering
    \includegraphics[width=1\linewidth]{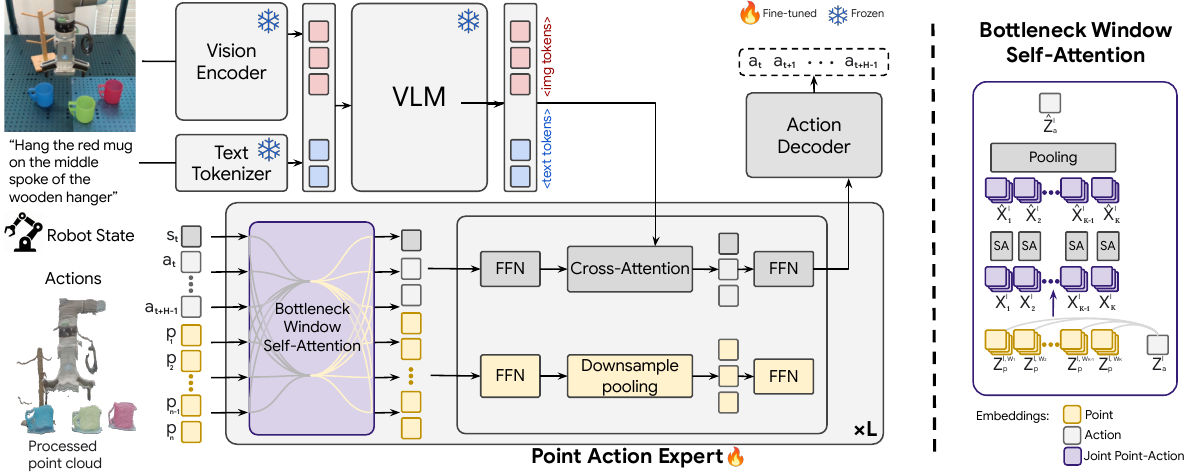}
    \caption{\textbf{(Left): PointACT Dual-Model Architecture. (Right): Bottleneck Window Self-Attention mechanism.} PointACT is a VLA model that equips a frozen pretrained VLM backbone with a point-cloud action expert for geometry-aware control. Language, images, robot state, and 3D point clouds are encoded into tokens, with point clouds producing multi-scale geometric features via a Point Transformer. These point tokens interact with action tokens through multi-scale point–action attention in the action expert, allowing both global structure and local geometry to guide action generation.
}
    \vspace{-1em}
    \label{fig:method}
\end{figure*}

%% file: sec/04_expr.tex
\section{Experiments}
\label{sec:expr}

\subsection{Implementation Details}
We utilize Qwen2.5-VL~\cite{bai2025qwen2_5vl} as the VLM backbone. 
To initialize from the pretrained PTv3 model~\cite{zhang2025concerto}, we use the same model configuration as PTv3-Large.  Specifically, there are 5 hierarchical stages with $(3, 3, 3, 12, 3)$ layers and feature dimensionality of $(64, 128, 256, 512, 768)$ per stage.
The PointACT module, including the hierarchical interaction blocks and the action head, comprises approximately 300M trainable parameters.
To maintain the knowledge of the VLM, we keep the VLM backbone frozen if not otherwise mentioned.
We crop the input point cloud using a workspace bounding box defined relative to the robot base frame~\cite{shridhar2023peract,chen2023polarnet}, and perform voxelization with size of 1cm. We set the maximum number of points as 4096.

Across our experimental benchmarks, the model is trained with a total batch size of 128 distributed across 2 NVIDIA H100 GPUs. We optimize the model for 20K to 50K gradient steps using the AdamW optimizer with a learning rate of $5 \times 10^{-5}$ and a cosine decay schedule. 
We apply standard point cloud augmentations, including random rotation around the gravity axis, alongside image-level augmentations for the VLM input during training.

\input{figs_tex/simulation_tasks}

\subsection{Evaluation Benchmarks}

\input{tables/libero}

We evaluate PointACT on two widely used simulation benchmarks LIBERO \cite{liu2023libero} and RLBench \cite{james2020rlbench}, to assess its performance in both short-horizon actions and long-horizon keypoint actions.
Figure~\ref{fig:simulation_tasks} presents some examples of the two benchmarks.

\noindent \textbf{LIBERO.}
We utilize four task suites from the LIBERO benchmark: \emph{LIBERO-Spatial}, \emph{LIBERO-Object}, \emph{LIBERO-Goal}, and \emph{LIBERO-Long}. 
Each suite contains 10 tasks with 50 expert demonstrations per task. 
Following the protocol in \cite{kim2025openvlaoft}, we filter unsuccessful demonstrations to ensure high-quality training data. 
For this benchmark, the action space consists of delta end-effector poses. Therefore, we employ the regression head with an action chunk size of $H=16$. 
During inference, we execute the first 8 steps before replanning.
For evaluation, we run 50 episodes per task following the standard protocol, resulting in 500 evaluation episodes per task suite. 
The supervised training and evaluation setup in the LIBERO benchmark is relatively easy, as test episodes differ from training data only by minor variations in object placement.
We report the average task success rate for each suite, using a binary score (1 for success, 0 otherwise) with no partial credit.

\noindent \textbf{RLBench.}
We use the same 10 RLBench tabletop manipulation tasks as HybridVLA~\cite{liu2025hybridvla}, including \emph{Close box}, \emph{Close laptop}, \emph{Toilet seat down},
\emph{Sweep to dustpan}, \emph{Close fridge}, \emph{Phone on base}, \emph{Take umbrella out}, \emph{Frame off hanger}, \emph{Wine at rack}, and \emph{Water plants}.
We train a unified multi-task policy using 100 demonstrations per task, and use the keypoint action space following prior work~\cite{liu2025hybridvla,shridhar2023peract,chen2023polarnet,guhur2023hiveformer,garcia2025gembench,pumacay2024colosseum}. 
The classification head is used for keypoint prediction with action chunk size $H=1$ since the action is already long-horizon.
An RRT-based motion planner is used to compute collision-free trajectories to the predicted keypoint.
Following the HybridVLA test setup, the testing episodes contain distinct object placements compared to the training episodes.
We evaluate each task over 100 episodes to ensure statistical reliability, as prior work has shown that motion planner in RLBench can introduce large execution variability~\cite{guhur2023hiveformer}.
The success rate is reported for each task.

For both LIBERO and RLBench benchmarks, we use a single front-view RGB image with $256 \times 256$ resolution. 
The point cloud $P_t$ is obtained from depth camera using known camera intrinsics and extrinsics.

\input{tables/hybridvla-10tasks}

\subsection{Comparison with State of the Art}

\noindent\textbf{LIBERO.}
Table~\ref{tab:libero_sota_cmpr} compares PointACT with prior VLA methods on the LIBERO benchmark. Direct comparison across VLAs is challenging because methods often use different training setups, data mixtures, and model configurations that are not always fully specified. 
For a fair comparison, we reproduce EO1~\cite{qu2025eo1} with our training pipeline. As a monolithic model, EO1 requires finetuning the VLM backbone (we keep the vision encoder frozen), resulting in more trainable parameters than PointACT. 
Even under this favorable setting, PointACT outperforms the reproduced EO1 by a large margin on Spatial and Long task suites, while achieving comparable performance on the Goal and Object suites. 
These results highlight the benefit of explicit 3D geometric information for spatially complex and temporally extended tasks. 
Additional controlled comparisons are provided in our ablation studies.

Our method outperforms most existing methods, though it performs 2\% worse than X-VLA~\cite{zheng2025xvla} and EO1~\cite{qu2025eo1}. However, results for these methods are not directly comparable, due to differences in experimental settings; specifically, those methods leverage pretraining on large-scale robot datasets and employ multi-view camera setups.

\noindent\textbf{RLBench.}
On RLBench, we report baseline results from the HybridVLA paper and reproduce EO-1 and ACT3D~\cite{gervet2023act3d} under our evaluation setup to ensure a fair comparison.
The compared methods, except for ACT3D, are image-based VLAs that do not use explicit 3D geometric inputs, whereas ACT3D is a state-of-the-art policy conditioned on both image and point cloud inputs.
Under the same evaluation protocol, PointACT achieves substantial performance gains over prior 2D- and 3D-based baselines across tasks. 

\input{figs_tex/failure_cases}

Figure~\ref{fig:failure_case_rlbench} illustrates representative failure modes across RLBench tasks, highlighting current limitations in spatial perception and long-horizon interaction.
\begin{itemize}
    \item \emph{Perceptual occlusion}: In the CloseFridge task, failure predominantly stems from partial views. This suggests multi-view integration or active perception is required to resolve occlusions.
    \item \emph{Limited failure recovery}: 
    Across several tasks, we observe a lack of reactive recovery once a trajectory deviates from the nominal path. If an initial prediction error leads to an unintended collision or a missed grasp, the model often fails to execute a corrective maneuver, instead continuing the pre-planned sequence. For example, in the Frame off Hanger task, the initial predicted pose is typically accurate; however, slight contact during the reach phase often shifts the object. The model does not recover from this failure but continues to perform the next step. This suggests that while 3D geometric features enhance initial precision, the model currently lacks the closed-loop temporal reasoning necessary to recover from high-level ``deadlock" states or physical perturbations.
    \item \emph{Tool-mediated interaction}: In the Sweep to Dustpan and Water Plant tasks, we observe inaccuracy in indirect object manipulation. While our PointACT accurately predicts primary contact points (e.g., grasping the broom), it struggles with the secondary spatial constraints of the tool itself. Specifically, inaccurate height estimation leads to insufficient contact between the broom and the dustpan, while collisions happen between the watering bottle and the plant. These cases indicate a need for better reasoning regarding tool-to-target spatial relationships.
\end{itemize}

\input{tables/3d_inject_ablation}

\subsection{Ablation Studies}

\noindent\textbf{3D injection architectures.}
We compare two strategies for integrating geometric information into the VLA framework as described in Section~\ref{sec:method_baselines}. 
The baselines include: (1)EO1~\cite{qu2025eo1}, a standard 2D VLA; (2) EO1 + Point, the monolithic 3D-aware architecture; (3) GR00T(arch), a dual-system VLA following GR00T~\cite{bjorck2025gr00t} architecture (using the same frozen VLM backbone as EO1 and without GR00T pretraining); (4)~GR00T(arch) + Point, which adds coarse-grained geometric tokens to the action expert; and (5) our proposed PointACT model with multi-scale point-action interaction.

Results are presented in Table~\ref{tab:3d_inject_ablation}.
For monolithic VLAs, injecting point cloud features does not achieve consistent improvements across benchmarks. It significantly decreases the performance on more challenging RLBench, where the training and testing episodes exhibit larger placement difference. This suggests that directly augmenting pretrained VLMs with 3D tokens does not effectively translate geometric information into improved action generation and may interfere with learned representations in VLM, and thus may require more pre-training with 3D integrated.
Adding point clouds to action experts improves performance more stably as shown in the dual-system VLAs. 
Compared to GR00T(arch)+Point which only utilizes coarse-grained point features from the last point encoder layer, the proposed PointACT achieves better results with fewer trainable parameters, demonstrating the effectiveness of fine-grained point–action interaction.

\noindent\textbf{Multi-scale point features.}
We further investigate the use of multi-scale point features in the more stable dual-system VLA setting.
As a straightforward baseline, we project features from different scales using separate MLPs, concatenate them into a single token sequence, and inject them into the action expert via cross-attention.
To keep the computational cost manageable despite the large number of point tokens, we sample a fixed number of tokens $K$ from each of the five scales in the PTv3 model.
As shown in Table~\ref{tab:ablation_vla2_multiscale_feature}, this naive multi-scale concatenation strategy does not improve performance for dual-system VLAs.
This result suggests that simply aggregating point features across scales is insufficient, and further demonstrates the necessity of our fine-grained multi-scale point-action interaction mechanism.

\begin{table}[t]
\centering
\caption{\textbf{Comparison of different strategies for incorporating multi-scale point features into dual-system VLAs.} We report the average success rate on RLBench-10Tasks. $K$ denotes the number of point tokens sampled from each scale.}
\label{tab:ablation_vla2_multiscale_feature}
\begin{tabular}{cccccc} \toprule
\multirow{3}{*}{} & \multirow{3}{*}{\begin{tabular}[c]{@{}c@{}}GR00T\\ (arch)\end{tabular}} & \multicolumn{3}{c}{GR00T(arch) + Point} & \multirow{3}{*}{\begin{tabular}[c]{@{}c@{}}PointAct\\ (Ours)\end{tabular}} \\
 & & \multirow{2}{*}{\begin{tabular}[c]{@{}c@{}}Final \\ layer\end{tabular}} & \multicolumn{2}{c}{Multi-scale} &  \\
 & & & K=64 & K=128 &  \\ \midrule
RLBench SR & 50.8 & 69.7 & 65.2 & 65.6 & \textbf{82.3} \\ \bottomrule
\end{tabular}
\vspace{-1.5em}
\end{table}

\noindent\textbf{Contribution of the 2D image modality.}
To isolate the importance of 2D visual context versus 3D geometric reasoning, we evaluate PointACT with and without image features. In the ``without image" variant, the model relies solely on language instructions from the VLM and the 3D point cloud. While 3D geometry provides strong spatial understanding, our results in Table~\ref{tab:image_embed_ablation} indicate that 2D visual features still provide critical semantic cues that complement the geometric representation.

\input{tables/image_embed_ablation}
\input{figs_tex/model_size_ablation}

\noindent\textbf{Model size of PointACT.}
We investigate the performance impact of the action expert's capacity by scaling the PointACT module. Following the PTv3 architectural configurations, we evaluate Small ($\sim$59M parameters), Base ($\sim$167M parameters), and Large ($\sim$314M parameters) variants. 
As shown in Figure~\ref{fig:model_size_ablation}, larger models slightly improve the performance.
Notably, the smallest PointACT action expert outperforms the baseline without a point cloud action expert shown in Table~\ref{tab:3d_inject_ablation}.

\noindent\textbf{PTv3 pretraining.}
We further measure the benefits from  pretrained PTv3D models in Figure~\ref{fig:model_size_ablation}.
There is a large domain gap between the pretraining dataset and the manipulation dataset, where the pretraining dataset consists of large-scale building-level scenes with many points that are coarsely sampled, while manipulation focuses on near areas with more densely sampled points.
Pretraining does not help much for small models, but provides greater benefits for larger models, potentially due to the increased optimization difficulty associated with higher-capacity architectures.
We will investigate pretraining on manipulation related environments~\cite{chen2024sugar} in future work.

\input{sec/05_real_robot_expr}

%% file: figs_tex/simulation_tasks.tex
\begin{figure}[t]
    \centering
    \begin{subfigure}{1\linewidth}
        \centering
        \includegraphics[width=\linewidth]{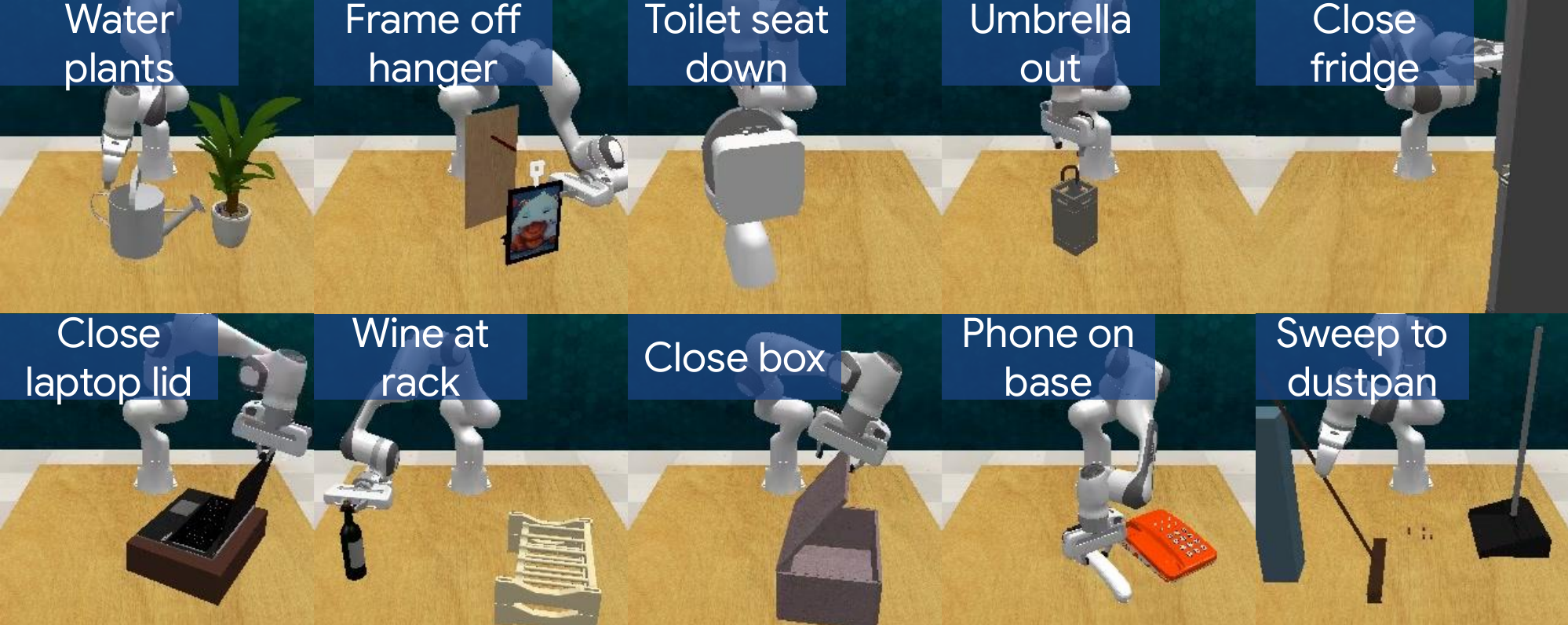}
        \vspace{-1.5em}
    \caption{RLBench 10 tasks.}
    \label{fig:rlbench_10_tasks}
    \end{subfigure}
    \hfill
    \begin{subfigure}{1\linewidth}
        \centering
        \includegraphics[width=\linewidth]{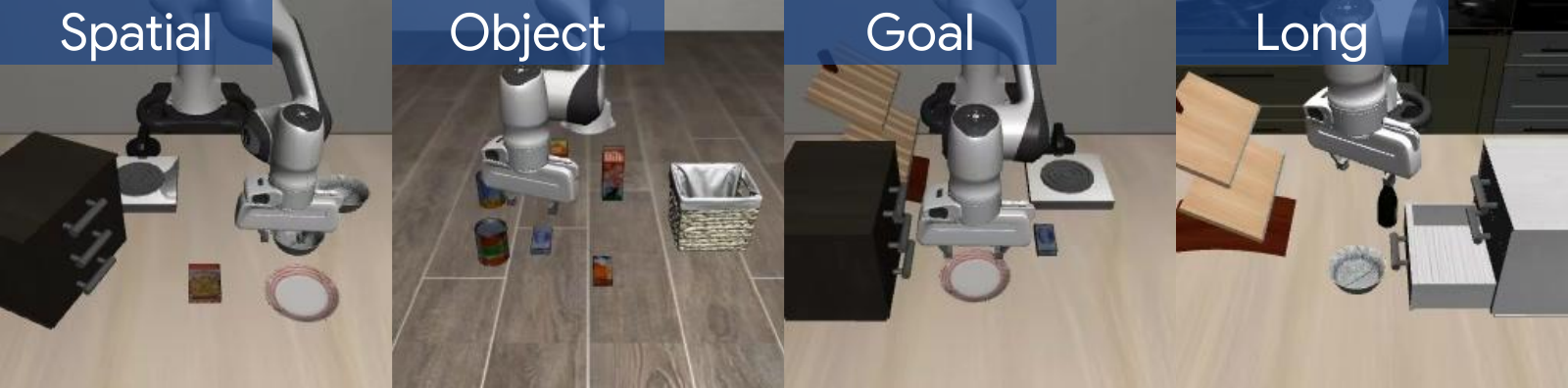}
        \vspace{-1.5em}
    \caption{LIBERO 4 task suites.}
    \label{fig:libero_task_suites}
    \end{subfigure}
    \caption{\textbf{Illustration of the simulated benchmarks.} 
    (a) Examples of tasks from the RLBench 10 tasks benchmark~\cite{liu2025hybridvla}, covering a diverse set of manipulation skills such as object placement, articulated object interaction and tool use. 
    (b) Representative tasks from the LIBERO benchmark~\cite{liu2023libero}, including spatial reasoning, object pick-and-place, goal-conditioned tasks, and long-horizon manipulation.}
    \label{fig:simulation_tasks}
    \vspace{-1em}
\end{figure}

%% file: tables/libero.tex
\begin{table*}[t]
\centering
\caption{\textbf{Success rate (\%) on the LIBERO benchmark.} ``Pretrained" denotes whether the model is fully pretrained, and ``train tasks" denotes the number of LIBERO tasks the model is fine-tuned on; some methods leverage all LIBERO tasks including LIBERO-90. We put a question mark if the setting is not clear from the original paper.}
\label{tab:libero_sota_cmpr}
\begin{tabular}{lcccccccccc} \toprule
 & \begin{tabular}[c]{@{}c@{}}Model \\ size\end{tabular} & \begin{tabular}[c]{@{}c@{}}Wrist\\ camera\end{tabular} & \begin{tabular}[c]{@{}c@{}}Pre-\\ trained\end{tabular} & \multicolumn{1}{c}{\begin{tabular}[c]{@{}c@{}}Frozen\\ VLM\end{tabular}} & \begin{tabular}[c]{@{}c@{}}Train\\ tasks\end{tabular} & Spatial & Object & Goal & Long & Avg. \\ \midrule
\rowcolor{softviolet} \multicolumn{11}{c}{2D VLAs} \\ %
TraceVLA~\cite{zheng2024tracevla} & 7B & ? & \checkmark & $\times$ & 40 & 84.6 & 85.2 & 75.1 & 54.1 & 74.8 \\
ThinkAct~\cite{huang2025thinkact} & 7B & ? & \checkmark & $\times$ & ? & 88.3 & 91.4 & 87.1 & 70.9 & 84.4 \\
Octo~\cite{ghosh2024octo} & 0.1B & $\times$ & \checkmark & $\times$ & 10 & 78.9 & 85.7 & 84.6 & 51.1 & 75.1 \\
OpenVLA~\cite{kim2025openvla} & 7B & $\times$ & \checkmark & $\times$ & 10 & 84.7 & 88.4 & 79.2 & 53.7 & 76.5 \\
OpenVLA-OFT~\cite{kim2025openvlaoft} & 7B & \checkmark & \checkmark & $\times$ & 10 & 97.6 & 98.4 & 97.9 & 94.5 & 97.1 \\
FLOWER~\cite{reuss2025flower} & 1B & \checkmark &  \checkmark & $\times$ & 10 & 97.1 & 96.7 & 95.6 & 93.5 & 95.7 \\
SmolVLA~\cite{shukor2025smolvla} & 2B & \checkmark & \checkmark & \checkmark & 40 & 93 & 94 & 91 & 77 & 88.8 \\
MolmoAct~\cite{lee2025molmoact} & 7B & \checkmark & \checkmark & $\times$ & 10 & 87 & 95.4 & 87.6 & 77.2 & 86.8 \\
GR00T-N1~\cite{bjorck2025gr00t} & 3B & \checkmark & \checkmark & ? & 100 & 94.4 & 97.6 & 93 & 90.6 & 93.9 \\
GR00T-N1.5~\cite{bjorck2025gr00t} & 3B & \checkmark & \checkmark & \checkmark & 10 & 92 & 92 & 86 & 76 & 86.5 \\
GR00T-N1.6~\cite{bjorck2025gr00t} & 3B & \checkmark & \checkmark & \checkmark & 10 & 97.7 & 98.5 & 97.5 & 94.4 & 97.0 \\
$\pi_0$~\cite{black2024pi_0} & 3B & \checkmark & \checkmark & $\times$ & 10 & 96.8 & 98.8 & 95.8 & 85.2 & 94.2 \\
$\pi_0$ + FAST~\cite{pertsch2025fast} & 3B & \checkmark & \checkmark & $\times$ & 10 & 96.4 & 96.8 & 88.6 & 60.2 & 85.5 \\
$\pi_{0.5}$~\cite{intelligence2504pi05} & 3B & \checkmark & \checkmark & $\times$ & 10 & 98.8 & 98.2 & 98.0 & 92.4 & 96.9 \\
X-VLA~\cite{zheng2025xvla} & 0.9B & \checkmark & \checkmark & $\times$ & ? & 98.2 & 98.6 & 97.8 & 97.6 & 98.1 \\
EO1~\cite{qu2025eo1} & 3B & \checkmark & \checkmark & $\times$ & 100 & 99.7 & 99.8 & 99.2 & 94.8 & \textbf{98.4} \\ 
\rowcolor{gray!15} EO1~\cite{qu2025eo1} (reproduced) & 3B & $\times$ & \checkmark & $\times$ & 10 & 91.8 & 98.2 & 96.6 & 85.6 & 93.1 \\ \midrule
\rowcolor{softviolet} \multicolumn{11}{c}{3D-aware VLAs} \\ %
SpatialVLA~\cite{qu2025spatialvla} & 4B & $\times$ & \checkmark & $\times$ & 10 & 88.2 & 89.9 & 78.6 & 55.5 & 78.1 \\
\rowcolor{gray!15} PointAct (Ours) & 3B & $\times$ & $\times$ & \checkmark & 10 & 97.4 & 99.6 & 96.2 & 90.6 & 96.0 \\
\bottomrule
\end{tabular}
\end{table*}

%% file: tables/hybridvla-10tasks.tex
\begin{table*}
\caption{\textbf{Success rate (\%) on RLBench 10 tasks.} 
We reproduce EO1~\cite{qu2025eo1} and ACT3D~\cite{gervet2023act3d}, and evaluate them alongside PointACT using 100 episodes per task.
The results for the other methods are directly taken from~\cite{liu2025hybridvla}, where each method is evaluated over 20 episodes.}
\label{tab:rlbench_sota_cmpr}
\begin{tabular}{lccccccccccc} \toprule
 & \begin{tabular}[c]{@{}c@{}}Close\\ box\end{tabular} & \begin{tabular}[c]{@{}c@{}}Close\\ laptop lid\end{tabular} & \begin{tabular}[c]{@{}c@{}}Toilet\\ seat down\end{tabular} & \begin{tabular}[c]{@{}c@{}}Sweep\\ to dustpan\end{tabular} & \begin{tabular}[c]{@{}c@{}}Close\\ fridge\end{tabular} & \begin{tabular}[c]{@{}c@{}}Phone\\ on base\end{tabular} & \begin{tabular}[c]{@{}c@{}}Umbrella\\ out\end{tabular} & \begin{tabular}[c]{@{}c@{}}Frame\\ off hanger\end{tabular} & \begin{tabular}[c]{@{}c@{}}Wine at\\ rack\end{tabular} & \begin{tabular}[c]{@{}c@{}}Water\\ plants\end{tabular} & Mean \\ \midrule
ManipLLM (7B)~\cite{li2024manipllm} & 50 & 80 & 40 & 20 & 80 & 35 & 10 & 25 & 15 & 20 & 38 \\
OpenVLA (7B)~\cite{kim2025openvla} & 65 & 40 & 75 & 60 & 80 & 20 & 35 & 15 & 10 & 10 & 41  \\
$\pi_{0}$~\cite{black2024pi_0} (2.6B) & 90 & 60 & {100} & 30 & 90 & 25 & 35 & {75} & 5 & 45 & 55 \\
CogACT (7B)~\cite{li2024cogact} & 80 & 85 & 90 & 65 & 90 & {50} & {60} & 35 & 25 & 25 & 60 \\
HybridVLA (7B)~\cite{liu2025hybridvla} & 85 & {95} & {100} & {90} & {100} & {50} & 50 & 70 & 50 & {50} & {74}  \\
EO1 (3B)~\cite{qu2025eo1} & 97 & 99 & 100 & 95 & 83 & 46 & 76 & 40 & 61 &  35 & 73.2 \\ 
ACT3D~\cite{gervet2023act3d} & 94 & 62 & 99 & 80 & 87 & 53 & 97 & 31 & 31 & 11 & 64.5 \\
\rowcolor{gray!15} PointACT (3B) & 91 & 99 & 96 & 59 & 81 & 99 & 99 & 69 & 90 & 40 & \textbf{82.3} \\

\bottomrule
\end{tabular}
\vspace{-1em}
\end{table*}

%% file: figs_tex/failure_cases.tex
\begin{figure}
    \centering
    \includegraphics[width=1\linewidth]{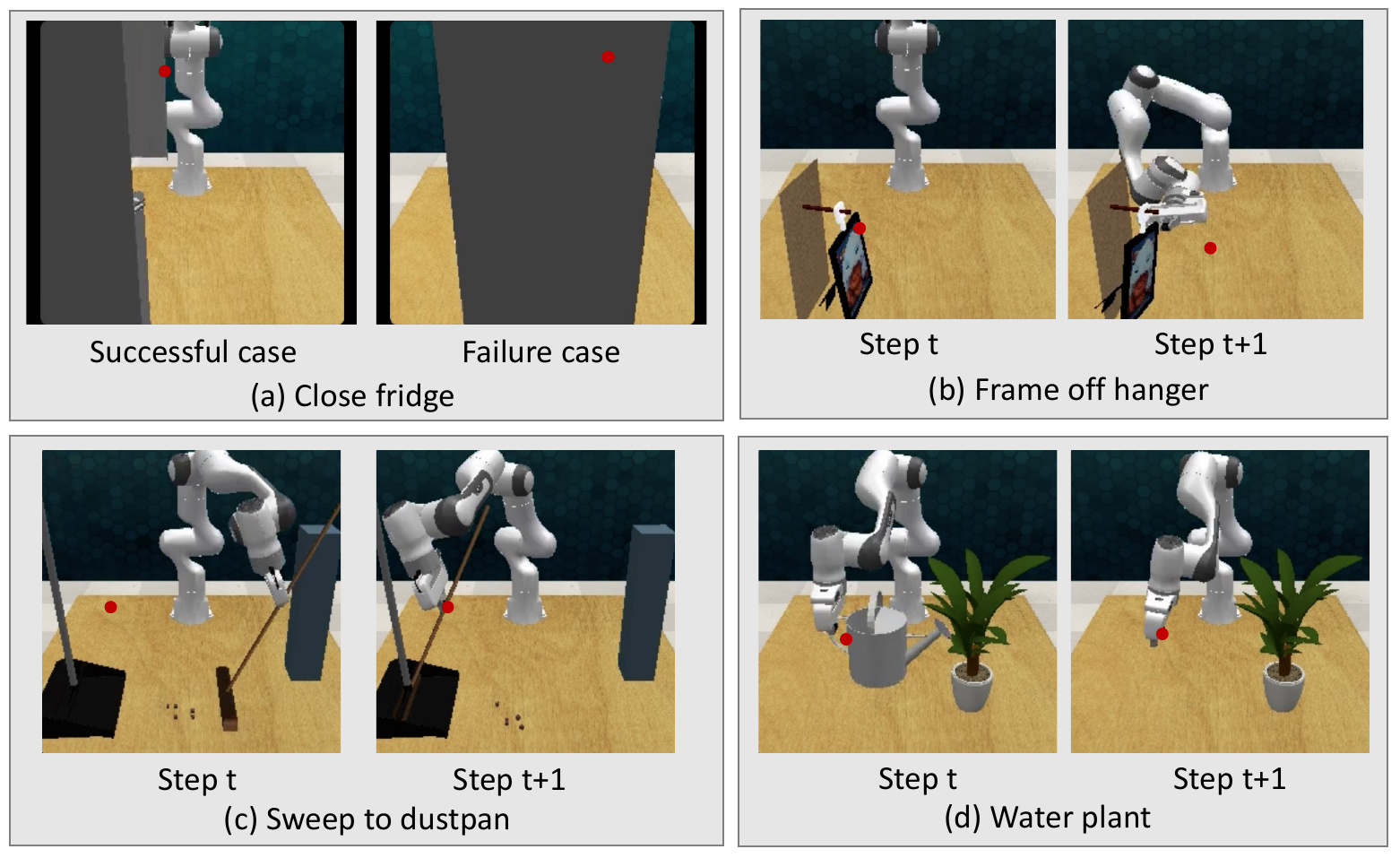}
    \caption{Failure cases in RLBench benchmark. The red dot denotes the position of the predicted next keypoint action projected in the front view image.}
    \label{fig:failure_case_rlbench}
\end{figure}

%% file: tables/3d_inject_ablation.tex
\begin{table}[t]
\caption{\textbf{Comparison of 3D integration strategies in VLAs.} We evaluate monolithic and dual-system VLAs with and without point cloud integration on LIBERO-Spatial and RLBench-10Tasks. For dual-system models, action experts are trained from scratch. GR00T(arch) uses the GR00T~\cite{bjorck2025gr00t} action expert architecture without pretrained weights.}
\label{tab:3d_inject_ablation}
\begin{tabular}{clccc} \toprule
 &  & \begin{tabular}[c]{@{}c@{}}Trainable\\ Params\end{tabular} & \begin{tabular}[c]{@{}c@{}}LIBERO\\ -Spatial\end{tabular} & \begin{tabular}[c]{@{}c@{}}RLBench\\ -10Tasks\end{tabular} \\ \midrule
\multirow{2}{*}{\begin{tabular}[c]{@{}c@{}}Monolithic\\  VLA\end{tabular}} & EO1 & 3B & 91.8 & 73.2 \\
 & EO1 + Point & 3B & 94.0 & 18.6 \\ \midrule
\multicolumn{1}{c}{\multirow{3}{*}{\begin{tabular}[c]{@{}c@{}}Dual-\\system\\ VLA\end{tabular}}} & GR00T(arch) & 1B & 87.0 &  50.8 \\
\multicolumn{1}{c}{} & GR00T(arch) + Point & 1B & 92.0 & 69.7 \\
\multicolumn{1}{c}{} & PointACT & 300M & \textbf{97.4} & \textbf{82.3} \\ \bottomrule
\end{tabular}
\vspace{-1em}
\end{table}

%% file: tables/image_embed_ablation.tex
\begin{table}[t]
\centering
\caption{Performance with and without conditioning on image embeddings from the frozen VLM.}
\label{tab:image_embed_ablation} 
\begin{tabular}{ccc} \toprule
Image conditioning & LIBERO-Spatial & RLBench-10Tasks \\ \midrule
$\times$ & 94.2 & 79.8 \\
\checkmark & \textbf{97.4} & \textbf{82.3} \\ \bottomrule
\end{tabular}
\end{table}

%% file: figs_tex/model_size_ablation.tex
\begin{figure}
    \centering
    \includegraphics[width=1\linewidth]{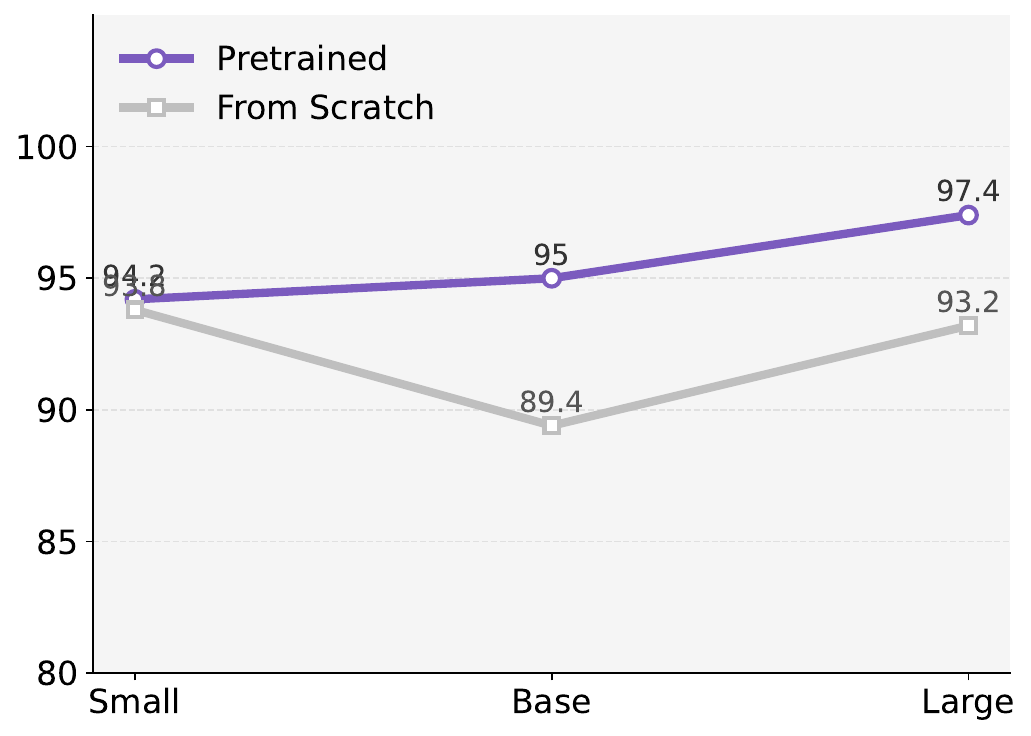}
    \caption{Performance on LIBERO-Spatial across different action expert sizes, with and without pretrained PTv3 weights~\cite{zhang2025concerto}.}
    \label{fig:model_size_ablation}
    \vspace{-1em}
\end{figure}

%% file: sec/05_real_robot_expr.tex
\section{Real World Experiments}

We evaluate our method on two distinct robotic platforms to assess performance under realistic conditions, including the SO-100 with microstep action control and the UR5 with keypoint-based actions. 

We compare our approach against two state-of-the-art VLAs: $\pi_0$~\cite{black2024pi_0} and GR00T-N1.5~\cite{bjorck2025gr00t}. 
We additionally compare with a state-of-the-art point cloud based policy 3DLotus~\cite{garcia2025gembench}, which is designed to predict keypoint actions, on the UR5 platform. 
Multi-task policies are trained to perform tasks following textual instructions.

\input{figs_tex/realrobot_setup}

\subsection{SO-100 Robot Platform}

\noindent\textbf{Experimental Setup.}
The SO-100 is a 6-DoF 3D-printed arm equipped with a parallel gripper. The workspace consists of a $50\text{cm} \times 50\text{cm}$ tabletop area monitored by a fixed front-view Intel RealSense D435 camera, see Figure~\ref{fig:so100_pointact_setup_reduced}. 
The observation space includes an RGB image, a point cloud obtained from depth maps and known camera parameters, and 6D joint positions.
The action space is the 6D absolute joint positions.

\begin{table}
\centering
\caption{\textbf{Performance on the SO-100 robot platform}. We report success rates over 10 trials, with partial scores shown in parentheses.}
\label{tab:so100_5tasks}
\begin{tabular}{lccc} \toprule
Task & $\pi_0$~\cite{black2024pi_0} & GR00T-N1.5~\cite{bjorck2025gr00t} & PointACT (Ours) \\  \midrule
Put Banana In Plate & \textbf{10/10} (10) & 8/10 (8) & \textbf{10/10} (10) \\
Put Sock In Drawer & 2/10 (5) & 5/10 (6.5) & \textbf{9/10} (9) \\
Open Microwave & 7/10 (7) & 5/10 (5) & \textbf{8/10} (8) \\  \bottomrule
\end{tabular}
\vspace{-1em}
\end{table}

\noindent\textbf{Tasks and Protocol.}
We evaluate performance across three manipulation tasks described in the following.  
We report success rate, but also a partial score defined for each tasks.

\begin{itemize}
    \item \textit{Put Banana In Plate:} Pick up a banana and place it on a blue plate. (Partial credit: 0.5 for grasp, 0.5 for placement)
    \item \textit{Put Sock In Drawer:} Pick up a sock amidst distractors, place it inside of a middle drawer, and finally close the drawer. (Partial credit: 0.5 for grasp, 0.25 for partial closure of the drawer; 0.5 for full closure)
    \item \textit{Open Microwave:} Open the microwave door to 90$^{\circ}$. (Partial credit: 0.5 for partially open, 1 for fully open)
\end{itemize}

We collect $50$ episodes per task for training using the leader-follower arms. 
Each episode involves randomized initial object placements.
Evaluation consists of 10 independent test episodes per task.
For fair comparison, all methods are tested using the same initial object positions.

\noindent\textbf{Baselines.}
We fine-tune baseline models on our collected SO-100 dataset following the standard configurations provided by the LeRobot library~\cite{cadene2024lerobot}.
For $\pi_0$, we use the official pretrained model weights and train the entire model parameters including the vision encoder with an action chunk size of 50.  
For GR00T-N1.5: We freeze the VLM backbone and fine-tune only the action expert with an action chunk size of 16.
Unlike the baselines, our model lacks massive robotic pre-training. Therefore, we initialize our PointACT from a LIBERO checkpoint and fine-tune the action expert (VLM is kept frozen) with an action chunk of 16.

\begin{table}[t]
\centering
\caption{\textbf{Performance on the UR5 robot platform.} We report success rates over 10 trials, with partial scores shown in parentheses.}
\label{tab:ur5_tasks}
\resizebox{\columnwidth}{!}{%
\begin{tabular}{lcccc} \toprule
Task & $\pi_0$ & GR00T-N1.5 & 3DLotus & PointACT \\ \midrule
Stack Yellow Cup Onto Pink Cup & 0/10 (2) & 0/10 (1.5) & 7/10 (7) & \textbf{7/10} (7.5) \\
Close Drawer & \textbf{9/10} (9) & \textbf{9/10} (9) & 2/10 (2) & 7/10 (8) \\
Put Grapes and Banana in Plates & 0/10 (2.25) & 0/10 (1.75) & 0/10 (3) & \textbf{4/10} (7) \\
\bottomrule
\end{tabular}%
}
\vspace{-1.5em}
\end{table}

\noindent\textbf{Results.}
Table~\ref{tab:so100_5tasks} shows the result with the SO-100 robot.
Despite the lack of extensive pre-training, our model achieves comparable or better performance than $\pi_0$ and GR00T. 

Compared to $\pi_0$, our approach performs similarly on Put Banana and Open Microwave tasks, but 
outperforms $\pi_0$ by a margin (2 versus 9 successful actions) on the more complex Put Sock task. 
While our approach correctly places the socks inside the drawer and manages to close it, 
$\pi_0$ often fails to place the sock fully inside the drawer, leading to collisions that prevent the drawer from closing. We hypothesize that this behavior is due to  missing 3D information.

Compared to GR00T, our approach performs consistently better.  We observe that GR00T often produces jittery actions, struggles with precise grasps, and can fail even on simple tasks such as Put Banana when objects appear in diverse poses. 
In contrast, PointACT generates noticeably smoother trajectories with reduced jitter and more accurate actions, benefiting from explicit 3D perception.

\subsection{UR5 Robot Platform}

\noindent\textbf{Experimental Setup.}
We use a 6-DoF UR5 robotic arm with a parallel RG6 gripper. The robot operates over a $1\,\text{m} \times 1\,\text{m}$ tabletop workspace observed by a fixed Orbbec Femto Mega RGB-D camera, see Figure \ref{fig:ur5_pointact_setup}.
Observations include an RGB image, a point cloud reconstructed from depth, and an 8D end-effector state (position, orientation, and gripper openness). 
Actions are specified as 8D absolute end-effector poses of the next keypoint, as collecting keypoint actions is easier with a joystick controller compared to continuous microstep actions.

\noindent\textbf{Tasks and Protocol.}
We evaluate performance across three manipulation tasks, and report success rate and a partial score. 
The tasks and partial success scores are defined as follows:

\begin{itemize}
    \item \textit{Stack Yellow Cup Onto Pink Cup:} Pick up a yellow cup amidst distractors, and stack it onto a pink cup. (Partial credit: 0.5 for grasp; 0.5 for placement)
    \item \textit{Close Drawer:} Move an open drawer (either top or middle-level drawer) to its closed position. (Partial credit: 0.5 for partially close; 1 for fully close)
    \item \textit{Put Grapes and Banana in Plates:} Pick up grapes amidst distractors, place it into a yellow plate, then pick up a banana amidst distractors, place it into a pink plate. (Partial credit: 0.25 per correct grasp or placement)
\end{itemize}

We collect $20$ episodes per task for training with a joystick controller, and evaluate on $10$ test episodes per task, with randomized initial object placements in each episode.

\noindent\textbf{Baselines.}
All models are fine-tuned on our collected UR5 dataset with action chunk size of 1 for keypoint action prediction. Following standard practice, the models are first pretrained on RLBench and subsequently fine-tuned on real-robot tasks using the pretrained checkpoints.

\noindent\textbf{Results.}
Table~\ref{tab:ur5_tasks} reports the results on the UR5 platform. 
The 3D-based policies outperform 2D-VLAs for all tasks except Close Drawer.
In this task, 
the drawer's transparency leads to noisy and incomplete point clouds. 
As a result, 3DLotus, which relies purely on geometric input, suffers a significant performance drop.
In contrast, our PointACT fuses RGB and point cloud cues, achieving substantially stronger results than 3DLotus. 
This result also highlights the importance of improving real-world point cloud quality.
On the cup-stacking task, our method achieves performance comparable to 3DLotus while substantially outperforming the 2D VLA baselines, $\pi_0$ and GR00T-N1.5. This task requires precise localization of the cup rim for successful grasping, which is challenging for 2D-only VLAs with limited robot data.
At the same time, the task relies less on semantic understanding, which narrows the performance gap between our method and the pure geometry-based 3DLotus baseline.
For the put fruit in plate task, both geometric precision and semantic understanding are important. PointACT achieves more accurate pre-grasp alignment, which we attribute to its fine-grained point cloud reasoning combined with semantic guidance from the VLM. Most failures occur during grasping, where small pose errors of the parallel gripper lead to consistent misses, particularly for the grapes.

Experimental results on the SO-100 and UR5 platforms demonstrate the superiority of our approach on real robots under noisy point clouds and actuation uncertainty. These results highlight the importance of joint 2D-3D reasoning with multi-scale point–action interactions.

%% file: figs_tex/realrobot_setup.tex
\begin{figure}[ht]
    \centering
    \begin{subfigure}{0.48\linewidth}
        \centering
        \includegraphics[width=\linewidth]{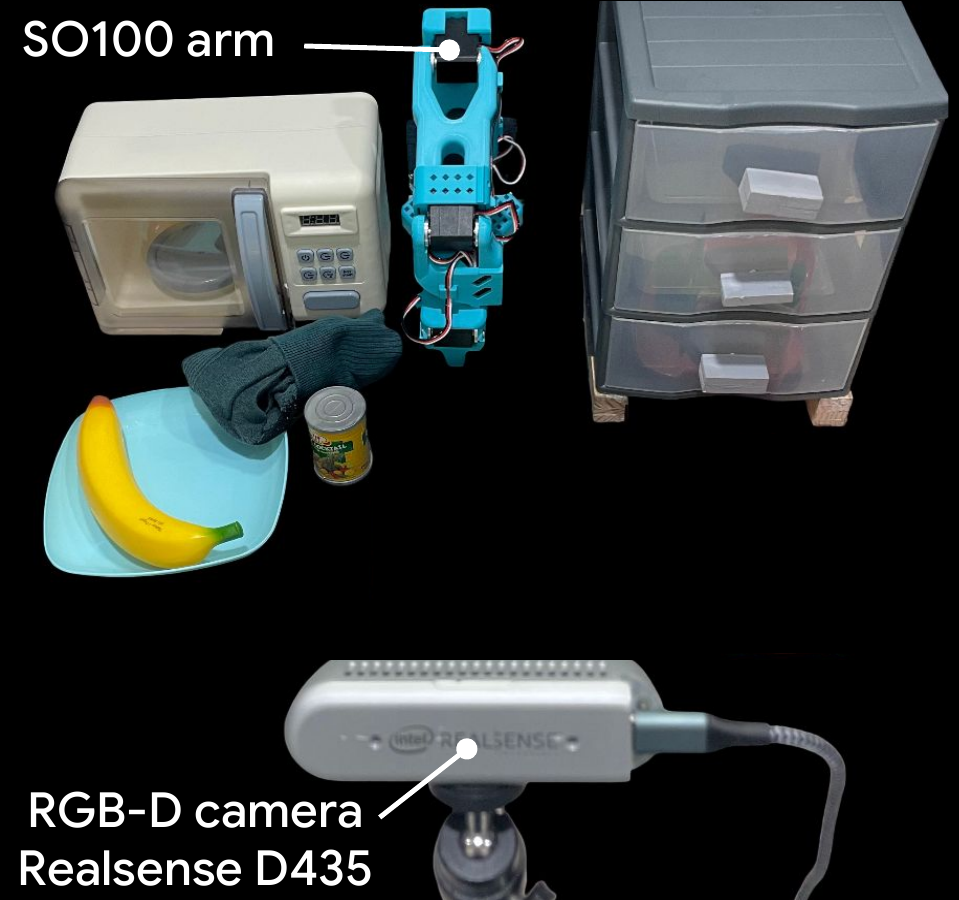}
        \caption{6-DoF SO100 arm with a front-view RGB-D camera.}
        \label{fig:so100_pointact_setup_reduced}
    \end{subfigure}
    \hfill
    \begin{subfigure}{0.48\linewidth}
        \centering
        \includegraphics[width=\linewidth]{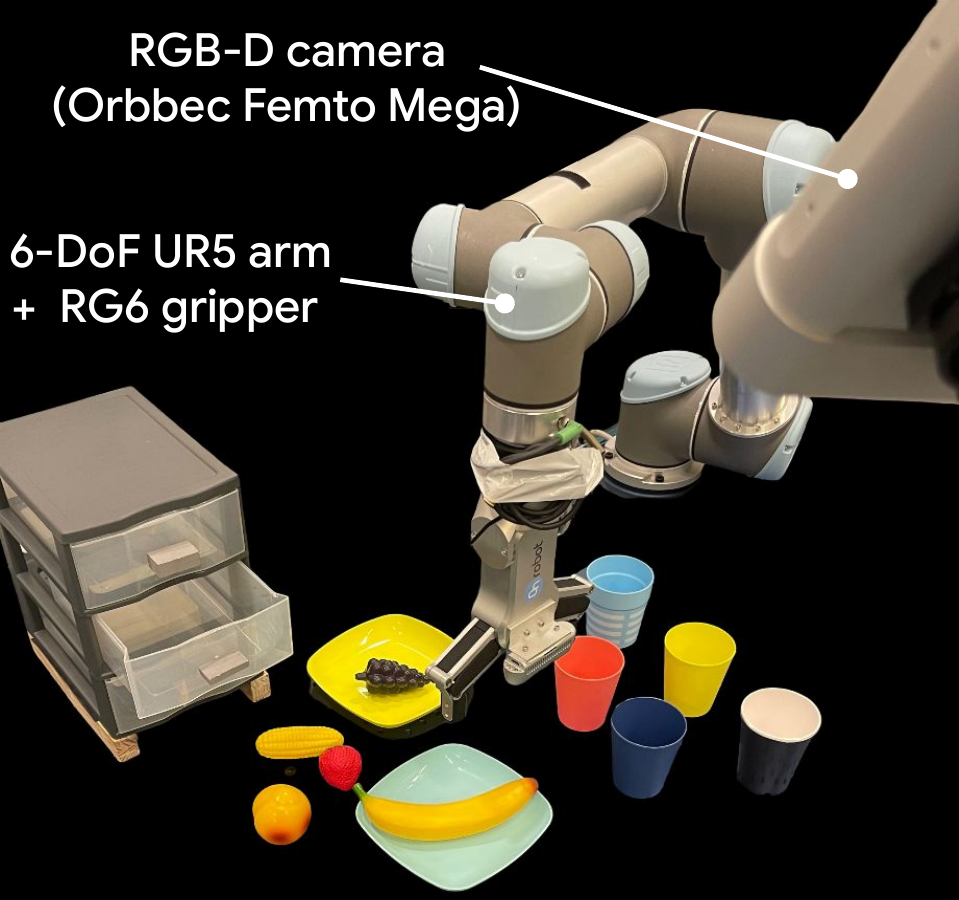}
        \caption{6-DoF UR5 arm with a front-view RGB-D camera.}
        \label{fig:ur5_pointact_setup}
    \end{subfigure}
    \caption{\textbf{SO100 and UR5 robot setups.}}
    \label{fig:real_robot_setups}
\end{figure}

%% file: sec/10_conclusion.tex
\section{Conclusion}
\label{sec:conclusion}

In this work, we presented PointACT, a 3D-powered Vision-Language-Action framework that integrates point clouds directly into action generation through a dedicated point-cloud action expert. Unlike prior approaches that attach 3D cues to image features or inject geometry into pretrained backbones, PointACT introduces multi-scale point-action interaction within the action decoder, enabling geometric information to continuously condition action tokens while preserving pretrained VLM representations. Our systematic study of 3D integration strategies shows that action-expert-level fusion with fine-grained point–action attention is significantly more effective than backbone-level injection or coarse geometric features.
Future work includes scaling point-based pretraining on large-scale robot datasets for manipulation-centric geometry, improving robustness under noisy point observations, and enhancing failure recovery abilities with 3D VLAs.

%% file: sec/12_appendix.tex
\appendix
We include efficiency analysis and additional experiments. 
Real-robot examples are provided in the project webpage.

\subsection{Efficiency analysis}
Table~\ref{tab:rlbench_mean_inferspeed} compares the average success rate and inference speed in frames per second (higher is faster) across 10 RLBench tasks for different VLA models.
EO1~\cite{qu2025eo1} and PointACT are evaluated by us on an NVIDIA H100 GPU.
The remaining results are taken from the HybridVLA paper~\cite{liu2025hybridvla}.
Due to differences in hardware and implementation details, the HybridVLA inference speed results are not directly comparable to our EO1 and PointACT evaluations and are included primarily for completeness.
Compared to EO1, our method achieves higher inference speed since it does not rely on flow matching, and also attains better overall performance due to its explicit 3D perception. 
Furthermore, we can observe that our approach outperforms the previous methods by a margin.

\begin{table}[h]
\centering
\caption{\textbf{Average success rate and inference speed on the RLBench 10-task benchmark.} 
Due to differences in hardware and implementation details, the results of the inference speed colored in gray are not directly comparable to our EO1 and PointACT evaluations and are included primarily for completeness.
}
\label{tab:rlbench_mean_inferspeed}
\begin{tabular}{lcc} \toprule
\textbf{Model} & \textbf{Success Rate} & \textbf{Inference Speed (Hz)} \\
\midrule
ManipLLM (7B)~\cite{li2024manipllm}        & 38 & {\color[HTML]{9B9B9B} 2.2}  \\
OpenVLA (7B)~\cite{kim2025openvla}        & 41 & {\color[HTML]{9B9B9B}6.3}  \\
$\pi_0$ (2.6B)~\cite{black2024pi_0}       & 55 & {\color[HTML]{9B9B9B}13.8} \\
CogACT (7B)~\cite{li2024cogact}         & 60 & {\color[HTML]{9B9B9B}9.8}  \\
HybridVLA (2.7B)~\cite{liu2025hybridvla}          & 58 & {\color[HTML]{9B9B9B}12.3} \\
HybridVLA (7B)~\cite{liu2025hybridvla}            & 74 & {\color[HTML]{9B9B9B}6.1}  \\
EO1 (3B)~\cite{qu2025eo1} & 73.2 & 3.8 \\
PointACT (3B) (Ours) & 82.3 & 6.7 \\ \bottomrule
\end{tabular}

\end{table}

\subsection{Additional Experimental Results}

\begin{figure}[h]
    \centering
    \includegraphics[width=1\linewidth]{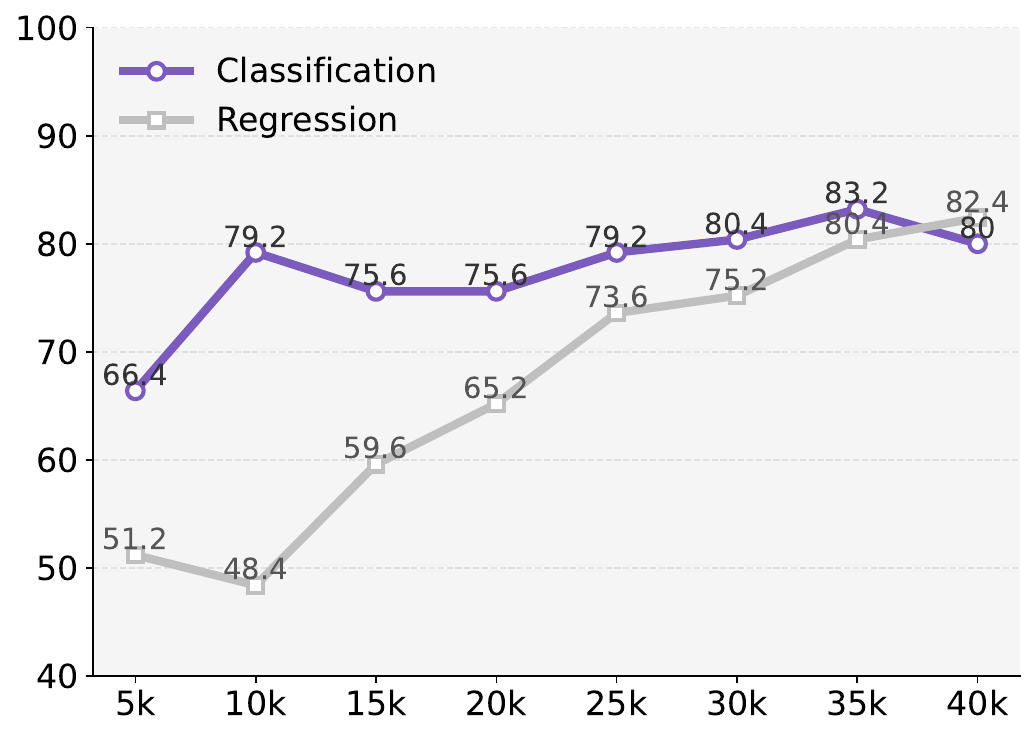}
    \caption{Comparison of classification and regression action prediction heads of our approach for varying training iterations. Results are reported on the RLBench 10-task benchmark and evaluated using 25 validation episodes per task.}
    \label{fig:cls_vs_reg_rlbench}
\end{figure}

\noindent\textbf{Classification head vs. Regression head.}
Although we adopt a classification head by default for keypoint-based action prediction, a regression head can also be used.
On the RLBench 10-task benchmark, the regression head achieves comparable final performance (82.4) to the classification head (82.3).
However, as shown in Figure~\ref{fig:cls_vs_reg_rlbench}, we observe that the classification head converges faster and yields better performance in the early training stage.
This is because discretized classification provides a more stable and stronger learning signal.

\noindent\textbf{VLA baselines with varying model sizes.}
For the dual-system VLA GR00T(arch), we vary the size of the action expert by changing its hidden dimension; the main paper uses a hidden size of 1536 in Table~III.
As shown in Table~\ref{tab:gr00t_size_rlbench}, increasing model capacity consistently improves performance for both models with and without point-cloud inputs.
Nevertheless, PointAct achieves better performance with substantially fewer parameters than the GR00T(arch) baseline, further demonstrating the effectiveness and parameter efficiency of our design.

\noindent \textbf{Performance variance across seeds.}
We evaluate on RLBench over three runs with different random seeds, obtaining $82.33_{\pm 0.65}$. 
This shows that the average is consistent with the value of 82.3 reported in the main paper and that the variance is insignificant.

\begin{table}[h]
\centering
\caption{Performance of GR00T(arch) with varying models sizes on RLBench-10Tasks.}
\label{tab:gr00t_size_rlbench}
\begin{tabular}{cccccc} \toprule
\multicolumn{2}{c}{Hidden size (GR00T)} & 768 & 1024 & 1536 & 2048 \\ \midrule
\multirow{2}{*}{\begin{tabular}[c]{@{}c@{}}No \\ point\end{tabular}} & \#Train params & $\sim$300M & $\sim$500M & $\sim$1B & $\sim$1.2B \\
 & RLBench SR & 43.0 & 45.1 & 50.8 & 54.2 \\ \midrule
\multirow{2}{*}{\begin{tabular}[c]{@{}c@{}}W/ \\ point\end{tabular}} & \#Train params & $\sim$500M & $\sim$700M & $\sim$1B & $\sim$1.3B \\
 & RLBench SR & 56.7 & 58.2 & 69.7 & 72.2 \\ \bottomrule
\end{tabular}
\end{table}

\begin{figure}[t]
    \centering
    
    \begin{subfigure}[h]{0.48\linewidth}
        \centering
        \includegraphics[width=\linewidth]{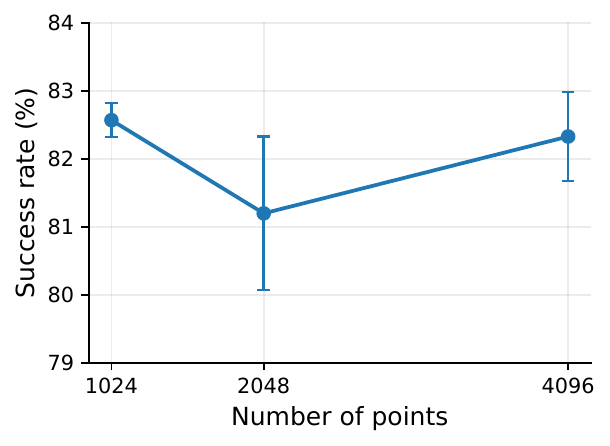}
    \end{subfigure}
    \hfill
    \begin{subfigure}[h]{0.48\linewidth}
        \centering
        \includegraphics[width=\linewidth]{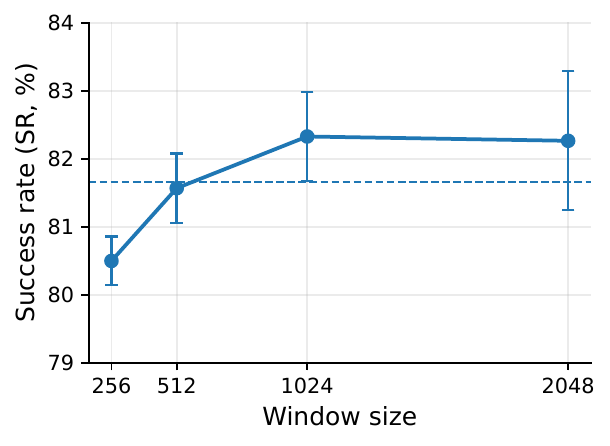}
    \end{subfigure}
    
    \caption{\textbf{Performance of PointAct with varying number of points and spatial window size on RLBench-10Tasks}.}
    \label{fig:pointact_ptv3_hyperparams}
\end{figure}

\noindent \textbf{Sensitivity analyses for voxelization and spatial windowing.}
We use a voxel size of 1cm, 4096 points, and a spatial window size of 1024 across all experiments.
The chosen voxel size follows common practice in prior 3D-based policies.
In Figure~\ref{fig:pointact_ptv3_hyperparams}, we vary the point cloud density and spatial window size. Our model remains robust to variations in point cloud density, and larger spatial window sizes consistently lead to improved performance.

